# Cybercars : Past, Present and Future of the Technology

**Michel Parent*, Arnaud de La Fortelle**
INRIA – Project IMARA
Domaine de Voluceau, Rocquencourt BP 105, 78153 Le Chesnay Cedex, France
Michel.parent@inria.fr

**ABSTRACT**— Automobile has become the dominant transport mode in the world in the last century. In order to meet a continuously growing demand for transport, one solution is to change the control approach for vehicle to full driving automation, which removes the driver from the control loop to improve efficiency and reduce accidents. Recent work shows that there are several realistic paths towards this deployment : driving assistance on passenger cars, automated commercial vehicles on dedicated infrastructures, and new forms of urban transport (car-sharing and cybercars). Cybercars have already been put into operation in Europe, and it seems that this approach could lead the way towards full automation on most urban, and later interurban infrastructures. The European project CyberCars has brought many improvements in the technology needed to operate cybercars over the last three years. A new, larger European project is now being prepared to carry this work further in order to meet more ambitious objectives in terms of safety and efficiency. This paper will present past and present technologies and will focus on the future developments.

## INTRODUCTION

Cybercars are fully automated road vehicles. A fleet of such vehicles forms a transportation system called CTS (Cybernetic Transportation System), for passengers or goods, operating in either a direct connection or an elaborate network, providing on-demand door-to-door transportation. The fleet of cars is under control of a central management system in order to distribute transportation requests efficiently and co-ordinate traffic in a particular environment. At the initial stages, cybercars are designed for short trips at low speed in an urban environment or in private grounds. In the long term, cybercars could also run autonomously at high speed on dedicated and protected tracks. With the development of the cybercar infrastructures, private cars with fully autonomous driving capabilities could also be allowed on these infrastructures while maintaining their manual modes on regular roads.

Cybercars are peoplemovers resembling PRT (Personal Rapid Transit). Their main advantage is their ability to operate at-grade on roads, which ensures they are cheaper and more flexible.

Although the concept was developed in the early 1990's, the fist CTS was put in operation at the end of 1997 in a long term parking at Schophol airport (Amsterdam). Since then, several other systems have been put in operation and many cities are now considering its implementation.

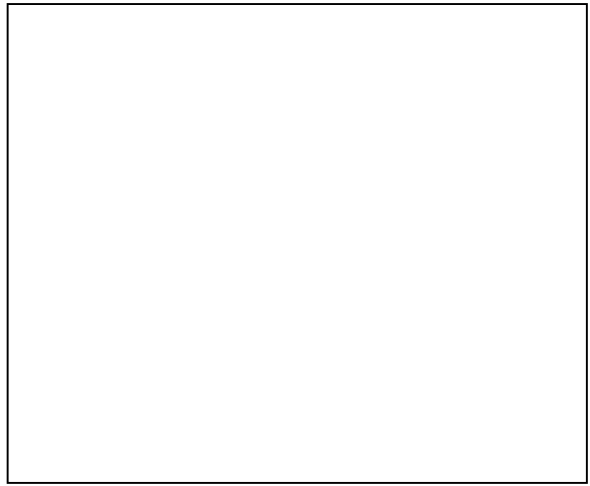

The future of the cybercars however lies in the integration of the cybercar features in regular cars (called dual-mode vehicles). These cars would then be allowed in restricted areas and on the dedicated infrastructures reserved for cybercars, and run in manual mode (with drivers assistance) on regular roads.

The technologies used for cybercars are similar to those found in drivers assistance techniques, and in particular use the same types of sensors and control. However, these techniques have to be pushed further because the vehicle is not under the responsibility of a human driver. However, on the other hand, the cybercars run at lower speed (they are restricted at the moment to urban environments) and in a more controlled environments. Further techniques, not available on standard vehicles at this time, must also be integrated in the cybercars for the fleet management and the interfaces with all sorts of users, who are not necessarily drivers.

We will now present the state of the art of the key cybercar technologies and the next developments which are foreseen.

**CONTROL**

Cybercars are precursors of drive-by-wire vehicles since acceleration, braking and steering are be controlled by computers. During the project, participants have developed new hardware for the safe implementation of these functions. However, the main focus has been on the development of safe software. To reach a high level of safety in a complex computer environment, often with distributed processing, a new tool developed by INRIA has been extensively used and validated by several partners. It is the SynDEx approach which allows the development and certification of distributed real time software (see http://www-rocq.inria.fr/syndex/).

The future of control system now relies in the development of redundant systems for the improvement of safety. The development of software in such distributed and redundant systems is still a difficult task and their certification is also a major difficulty.

**OBSTACLE AVOIDANCE**

Obstacle avoidance is the main difficulty in the deployment of cybercars. Considerable research work has been carried out in this domain by the partners. Now available on industrial vehicles are systems based on scanning laser rangefinders complemented by ultra-sounds and sensing bumpers. These sensors are associated with advanced control software to anticipate potential collision while eliminating obstacles which are not on the path of the vehicles.

Other collision avoidance techniques based on radar and on vision have been researched. They are not yet certified but offer great promise for lowering the cost and improving the performance. These researches are conducted in close cooperation with the automobile industry which is looking for similar devices for avoiding vulnerable users in urban environments.

For the vision, there is a trend to use hierachical methods, with hardware to deal with the massive amount of low-level operations, then sending higher level objects to various routines. E.g. Genetic Algorithms (as the Fly Algorithm) can be implemented in hard and can very quickly give some hints of the presence of pedestrians in the path of the vehicle and that information is used to launch refined, computer demanding methods in these particular areas.

**PLATOONING**

Platooning techniques are needed for the operation of several vehicles closely spaced. The first vehicle of this platoon may or may not be automatic depending on the application. Two techniques have been developed in the project. On relies on the scanning laser sensor used for obstacle avoidance and the other is based on the development of a linear camera using low cost components. Both approaches give good results but the linear camera has the potential for very low gap and high speed operation.

The next generation of platoons will clearly communicate in order to solve problems such as inserting a car into the platoon, splitting the platoon or to be able to manage intersection of two platoons (e.g. at intersections). This will also help with the *stability* problem of platoons.

**LOCALISATION AND NAVIGATION**

The first automated vehicles used an infrastructure-based approach with electric wires or transponders. During the project, Frog has further developed their technique based on dead-reckoning associated with relocalisation on magnets widely spaced and hence implemented at low cost. This technique allows for fine tuning the exact path of the vehicles and is available on the ParkShuttle II and requires less magnets on the road than previously.

Other techniques based on localisation by laser or natural features in the environment or on vision have been demonstrated. These techniques which require no modification of the environment are still to be industrialised. Advanced techniques for path generation in complex and dynamic environments have also been explored successfully.

A low cost solution being developed is to use the same architecture as the obstacle detection for localization. Extracting higher-level feature allows to reduce drastically the amount of data stored in the GIS (Geographical Information System) with respect to current low-level maps. Then a 2D path is computed, taking into account all the obstacles seen or transmitted by the surrounding sensors: the nearby cybercars share their information in order to enhance the range and the precision of detection of the surrounding.

**FLEET MANAGEMENT**

During the project, the industrial companies have developed management software based on a centralised system and communications. These systems now offer a very flexible operation and can implement a demand responsive transportation system with minimum waiting times and a low number of vehicles.

At the research level, new techniques have been developed for the optimisation of large scale systems, including hierarchical control. There is an overall fleet control for navigation that ensures also good redistribution of cybercars so that offer and demand coincide. At a local level, intersections manage incoming cybercars so that throughput and safety are optimized. At the cybercar level the onboard control deals with the trajectory and obstacle avoidance.

**COMMUNICATIONS**

Good communication between the vehicles and the infrastructure and between infrastructure and the users is essential for any good transportation system. In the case of cybercars where the vehicles are run according to demand, this is even more essential. During the project, various communication schemes have been used and are now operational on various systems: GSM and GPRS mostly for communicating with the users through their mobile phones, and Wi-Fi (IEEE 802.11) for the communication between vehicles and infrastructure. High bandwidth communication is needed in case of transfer of images, for example for remote control of the vehicles.

Mobile ad hoc networks are now being improved to take into account the particular mobility of cybercars (e.g. adapted versions of the OLSR protocol). They are very well suited to cybercars sytems since they always offer a sufficient density. Next steps are to demonstrate this technology in large system and to deal with the handover between different communication means (2G, 3G, WiFi, satellite...).

**ENERGY**

Cybercars offer the unique opportunity to turn away from internal combustion engines and inherent local pollution and noise in the cities. All the cybercars available now run on batteries and electric motors. Due to the low energy capacity of the batteries, the management of the energy is crucial for an efficient operation of the system. Various optimisation algorithms have been developed in the project for the optimum battery capacity and recharging strategy. Also, techniques for automatic recharging and for energy transfer through induction have been developed and tested.

**HMI**

Human machine interfaces (HMI) are also one of the key elements for the ease of use of the system and hence for its acceptance. Various developments have been done in the project to work on simple but powerful interfaces inside or outside of the vehicle. It has been accepted now that the most convenient way to request the vehicles is either through simple call buttons (such as for elevators) or, when this is not possible due to a very large number of pick-up points, through a mobile phone. More advanced interfaces have also been explored in the context of another European Project: OZONE, which develops the concept of "ambient intelligence". The cybercars are used in this project as a test case.

## CERTIFICATION

Safety and reliability are major issues for the introduction of new systems like CyberCars. Traditional vehicles that use the public roads have to meet a large number of requirements, laid down in European standards and regulations. For new systems like CTS, such standards do not yet exist and the existing standards for other road vehicles are not always suitable. Therefore recommendations for certification standards were established as a part of the project. These standards will help manufacturers and operators of cybercars to assess the safety of their systems and to establish how safe a system should be.

## AKNOWLEDGMENTS

This work on the development of cybercars technologies was partly financed by the European Commission Programme IST.

## REFERENCES

See articles and reports at : www.cybercars.org